\documentclass[review]{cvpr}
\usepackage{times}
\usepackage{epsfig}
\usepackage{graphicx}
\usepackage{amsmath}
\usepackage{amssymb}
\usepackage{rotating}
\usepackage{multirow}
\newcommand*{\Scale}[2][4]{\scalebox{#1}{$#2$}}%

\def\xx{\mathbf{x}}
\def\ff{\mathbf{f}}
\def\bb{\mathbf{b}}

\def\uu{\mathbf{u}}

\def\wn{\mathbf{\tilde{w}}}
\def\wni{\tilde{w}}
\def\WWn{\mathbf{\tilde{W}}}

\def\bb{\mathbf{b}}

\def\R{{\rm I\!R}}   

\def\cc{\mathbf{s}}
\def\ww{\mathbf{w}}

\def\xgamma{\boldsymbol{\gamma}}
\def\gamman{\tilde{{\boldsymbol{\gamma}}}}
\def\gammani{\tilde{\gamma}}

\usepackage[pagebackref=true,breaklinks=true,colorlinks,bookmarks=false]{hyperref}

\def\cvprPaperID{1065} 
\def\confYear{CVPR 2021}

\begin{document}

\title{Deep Compact Polyhedral Conic Classifier for \\ Open and Closed Set Recognition}





\author{\parbox{16cm}{\centering
    {\large Hakan Cevikalp$^1$, \hspace{0.2cm} Bedirhan Uzun$^1$, \hspace{0.2cm} Okan K\"op\"ukl\"u$^2$, \hspace{0.2cm} Gurkan Ozturk$^3$}\\
    {\normalsize
    \vspace{0.2cm}
    $^1$ Eskisehir Osmangazi University \\
    $^2$ Technical University of Munich \\
    $^3$ Eskisehir Technical University}}
}

\maketitle

\begin{abstract}
   In this paper, we propose a new deep neural network classifier that simultaneously maximizes the inter-class separation and minimizes the intra-class variation by using the polyhedral conic classification function. The proposed method has one loss term that allows the margin maximization to maximize the inter-class separation and another loss term that controls the compactness of the class acceptance regions. Our proposed method has a nice geometric interpretation using polyhedral conic function geometry. We tested the proposed method on various visual classification problems including closed/open set recognition and anomaly detection. The experimental results show that the proposed method typically outperforms other state-of-the-art methods, and becomes a better choice compared to other tested methods especially for open set recognition type problems.  
\end{abstract}

\section{Introduction}

Recently deep convolutional neural network classifiers have significantly dominated the computer vision field by yielding the state-of-the-art performance in many classification tasks such as visual object, face, scene, and action classification and detection problems. The soft-max loss (more precisely, cross entropy loss on probabilities obtained by applying soft-max function to the output logits of the deep neural network) is the most commonly used classification loss function in the classical convolutional neural networks. However, many studies \cite{R1,R2,R3,R10,R12} demonstrated that soft-max loss works well for closed set recognition problems where the test samples come only from the classes seen during training. However, soft-max typically fails for open set recognition problems, the settings that allow the test samples coming from the unknown classes \cite{R33}. This is partly due to the two important reasons: classical soft-max does not enforce a large-margin between classes and there is no mechanism to control the intra-class variations. In open set recognition problems, novel classes (ones not seen during training) may occur at test time, and classifiers are expected to reject the test samples that do not appear to belong to any of the known training classes. Therefore, compactly clustered and well separated acceptance regions estimated for classes improve accuracies in such open set recognition problems. Here, acceptance region of a class indicates the bounding region where a test sample is assigned to that class label if its feature representation lies in that region and rejected if its feature vector is outside the region.  

To return more compact class acceptance regions, \cite{R4,R5} use contrastive loss (CL). Contrastive loss minimizes the Euclidean distance of the positive sample pairs and penalizes the negative pairs that have a distance smaller than a given threshold. In a similar manner, \cite{R6,R7,R8,R9} employ triplet loss function that used a positive sample, a negative sample and an anchor. An anchor is also a positive sample and the goal is simultaneously to minimize the distance between the positive sample and the anchor and to maximize the distance between the anchor and the negative sample. Although methods using both contrastive and triplet loss functions return compact decision boundaries, the number of sample pairs or triplets grows quadratically compared to the total number of samples, which results in slow convergence and instability. To overcome this limitation, a careful sampling/mining of data is required, and the accuracy largely depends on this mining strategy. Recent studies \cite{R8,R9,R40} also reveal that large batch sizes during learning and selecting the same number of samples per class are necessary for good results.

Wen et al. \cite{R1,R2} combined the soft-max loss function with the center loss for face recognition. Center loss reduces the intra-class variations by minimizing the distances between the individual class samples and their corresponding class centers. The resulting method significantly improves the accuracies over the method using soft-max alone in the context of face recognition, which shows the importance of compact class decision boundaries. Despite its good accuracies, this method does not explicitly attempt to maximize the inter-class separability and relies on soft-max for this goal. 
Zhang et al. \cite{R10} combined the range loss with soft-max loss to overcome the limitations of the contrastive loss. The range loss consists of two terms that simultaneously minimize the intra-class variations and maximize the inter-class separation. 
However, this method is very complex since one has to carefully tune two weights to balance 3 different loss terms. Furthermore, it only penalized one center pair in each batch to maximize the inter-class separation. As pointed out in \cite{R11}, this strategy is not comprehensive since more center pairs may have margins smaller than a designated threshold. Deng et al. \cite{R12} introduced a method combining soft-max loss function with the marginal loss to create compact and well separated classes in Euclidean space. Marginal loss enforces the distances between sample pairs from different classes to be larger than a selected threshold while enforcing the distances between sample pairs coming from the same classes to be smaller than the selected threshold. Wei et al. \cite{R11} combined soft-max loss  and center loss functions with the minimum margin loss where the minimum margin loss enforces all class center pairs to have a distance larger than a specified threshold. This method is also complex since there are many parameters that must be carefully set by the user. 
Support Vector Data Description (SVDD) method of \cite{R35} aims to find a compact hypersphere that includes the majority of the positive class samples for anomaly detection. This method is extended in \cite{R34} by using deep neural networks, where the proposed classifier trains a deep neural network that minimizes the volume of hypersphere that encloses the deep network representations of the data. However this method has limitations in the sense that the hypersphere center is fixed to a certain vector at the beginning and it is not updated as in center loss method. Also, it requires to use only ReLU activation functions to avoid trivial solutions.

Liu et al. \cite{R14,R15} proposed the SphereFace method which uses the angular soft-max (A-softmax) loss that enables to learn angularly discriminative features. To this end, the proposed method projects the original Euclidean space of features to an angular space, and introduces a multiplicative angular margin to maximize inter-class separation. However, the SphereFace  method has limitations in the sense that the decision margins change for each class. As a result, some inter-class features have a larger margin while others have a smaller margin, which reduces the discriminating power \cite{R16}. Zhao et al. \cite{R17} proposed the RegularFace method in which A-softmax term is combined with exclusive regularization term to maximize the inter-class separation. Wan et al. \cite{R16} introduced the CosFace method which imposes an additive angular margin on the learned features. To this end, they normalize both features and learned weight vectors to remove radial variations and then introduce an additive margin term $m$ to maximize the decision margin in the angular space. 
A similar method called ArcFace is introduced in \cite{R18}, where an additive angular margin is added to the target angle to maximize the separation in angular space. It should be noted all these aforementioned methods, which maximize the margin in the angular space rather than the Euclidean space, are proposed for face recognition. These methods work well for face recognition since face class samples in specific classes can be approximated by using linear/affine spaces, and the similarities can be measured well by using the angles between sample vectors in such cases. Linear subspace approximation will work as long as the number of the features are much larger than the number of class specific samples which holds for many face recognition problems. However, for many general classification problems, the training set size is much larger compared to the dimensionality of the learned features and therefore these methods cannot be generalized to the classification applications other than face recognition as demonstrated in our experiments.

Regarding open set recognition, in addition to the methods using hand crafted features such as \cite{R36,R37}, recent methods \cite{R38,R39} use deep neural networks for such settings. Neal et al. \cite{R38} introduce a method that uses an encoder-decoder GAN (Generative Adversarial Network) architecture to generate synthetic open set examples. Oza and Patel \cite{R39} propose an open set recognition algorithm using class conditioned auto-encoder with new training and testing methodologies. 

\medskip\noindent{\bf Contributions:} In this paper, we propose a general deep neural network classifier that returns compact and well separated acceptance regions for each class. The proposed methodology allows to use a margin-based cost function to maximize the inter-class separation and there is a mechanism that enables to control the compactness of the class regions to minimize the intra-class variations. As opposed to the methods combining soft-max loss function with other loss terms, our deep neural network gracefully integrates intra-class compactness and inter-class separation goals into a single unified framework. Moreover, in contrast to the some of the aforementioned methods above, the proposed method has a nice geometric intuition that helps to interpret the resulting class acceptance regions.

Our proposed method returns a compact polyhedral acceptance region for each class. The samples coming from other classes can be easily rejected based on the distances to the resulting acceptance regions. This makes the proposed method suitable especially for open set recognition and anomaly detection problems. Our proposed method does not need to generate synthetic samples and it significantly outperforms state-of-the-art open set recognition methods such as \cite{R38,R41} relying on synthetically created open set examples.

\section{Method}
This paper introduces a deep neural network classifier that returns well separated compact class decision boundaries. To this end, we propose a new loss function for CNN networks. In the proposed method, inter-class separation is realized by using the hinge loss and we have another loss term to adjust the compactness of the acceptance regions of the classes and class features. We build our method based on polyhedral conic classifiers introduced in \cite{R20,R19}. Therefore, we first briefly overview polyhedral conic functions and then introduce our new deep neural network classifier.
\subsection{Preliminaries}
In polyhderal conic classification methods, each class is approximated with a polyhedral conic region constructed by using polyhedral conic functions (PCFs). PCFs were first introduced in \cite{R20} to separate two arbitrary finite disjoint sets, and its extended version, extended polyhedral conic functions (EPCFs), have been proposed in \cite{R19} for visual object detection, closed and open set recognition. PCFs and EPCFs can be defined as follows:
\begin{align}
 \!\! f_{\ww,\gamma,\cc,b}(\xx)&=\ww^\top(\xx-\cc)+\gamma\left\|\xx-\cc \right\|_1 -b, &\textrm{(PCF)}\!\!\label{eq:con_func} \\
  \!\! f_{\ww,\xgamma,\cc,b}(\xx)&=\ww^\top(\xx-\cc)+\xgamma^\top|\xx-\cc| -b, &\textrm{(EPCF)}\!\!\label{eq:econ_func}
\end{align}
where $\xx \in \R^d$ is a test point, $\cc\in\R^d$ is the cone vertex, $\ww \in \R^d$ is a weight vector and $b$ is an offset. For PCF, $\left\|\uu\right\|_1=\sum_{i=1}^d|u_i|$ denotes the vector L1 norm and $\gamma$ is a corresponding weight, while for EPCF, $\left|\uu\right|=\left(|u_1|,\ldots,|u_d|\right)^\top$  denotes the component-wise modulus and $\xgamma$ is a corresponding weight vector. 

\begin{figure*}[tbh]
\centering
\begin{tabular}[h]{@{}ccc@{}}
\includegraphics[width=45mm]{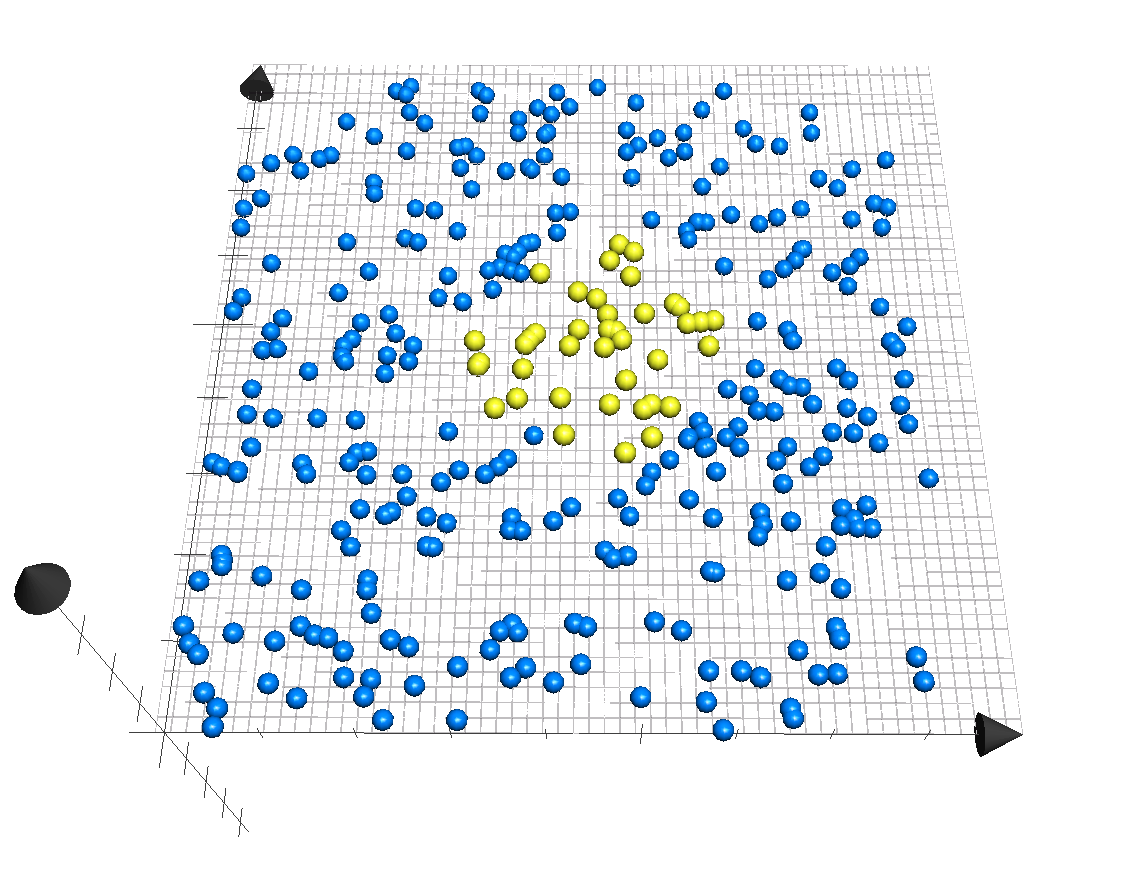} &
\includegraphics[width=45mm]{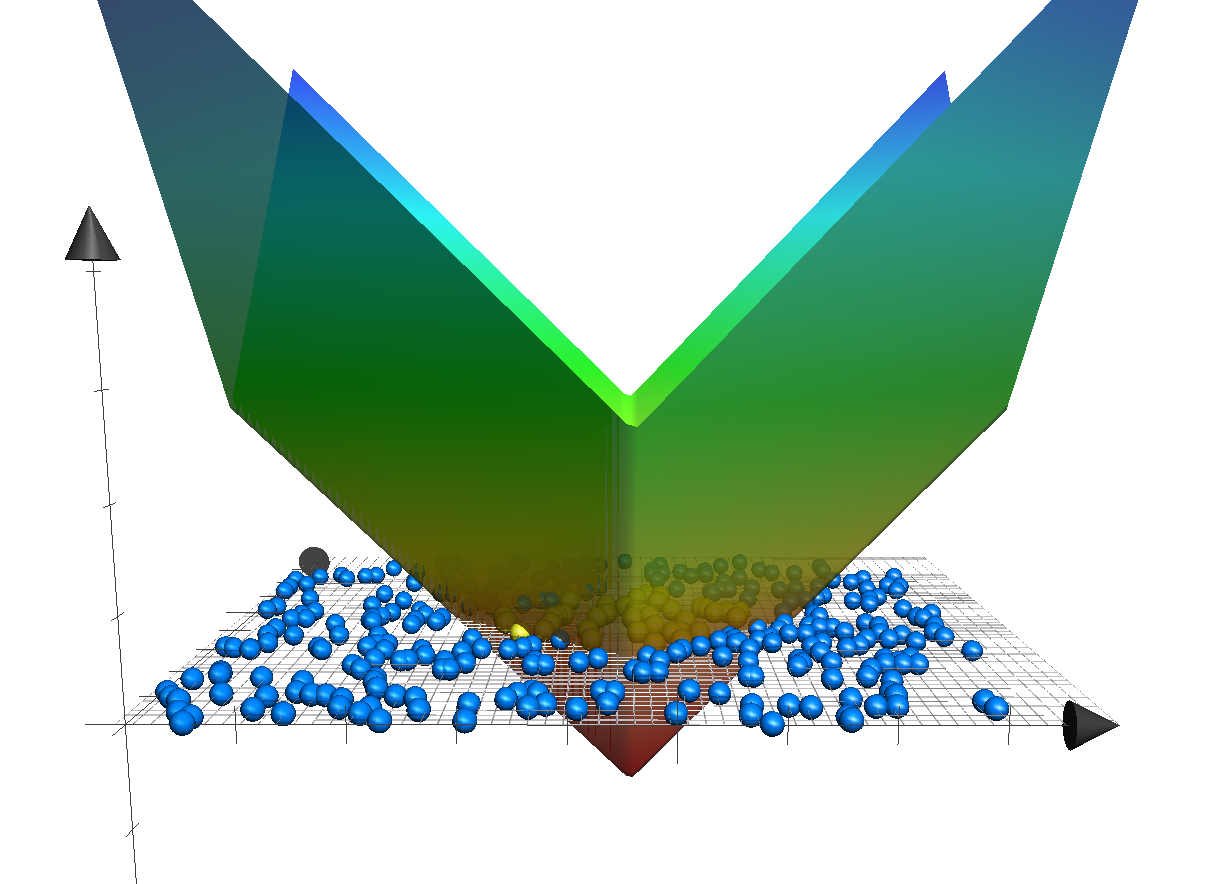}  & \includegraphics[width=45mm]{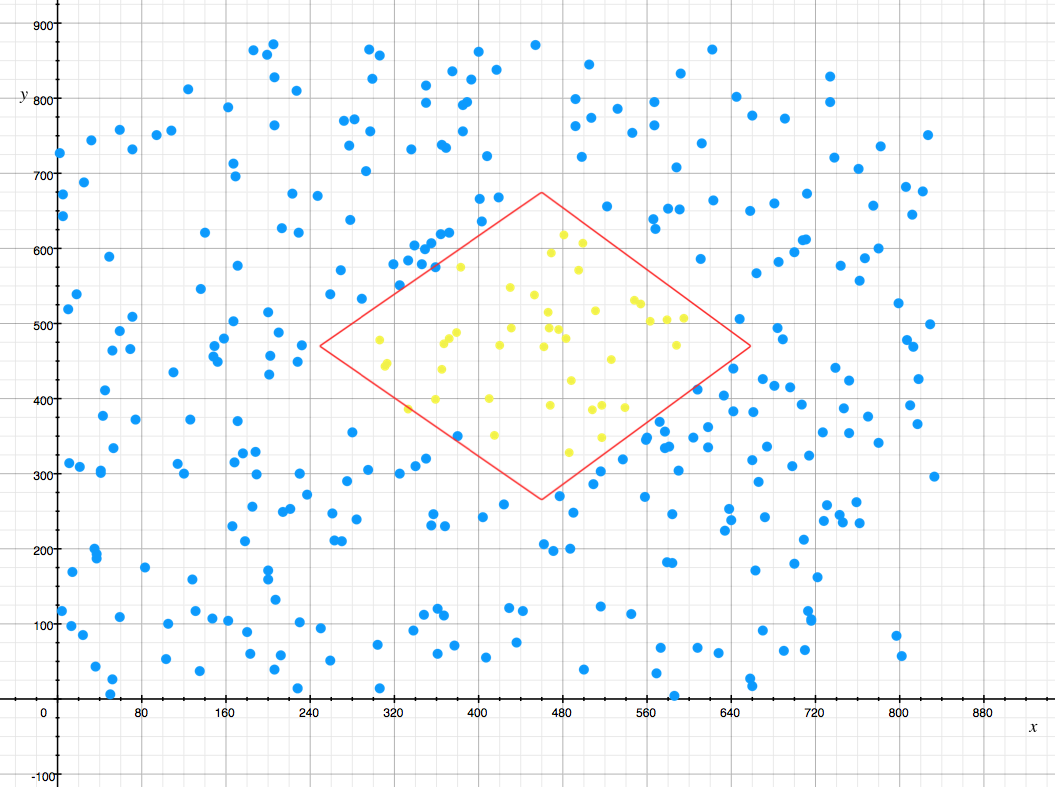} \\
(a) & (b) & (c) 
\end{tabular}
\caption{Visualization of polyhedral conic classifier (PCC) for 2D synthetic data: The positive acceptance regions are “kite-like” octahedroids containing the points for which a linear hyperplane lies above an $L_1$ cone.(a): 2D positive (yellow) and negative (blue) samples, (b) view of positive-class acceptance region in 3D, (c): Resulting “kite-like”  acceptance region in 2D space. (Figure is adopted from \cite{R19})}
\label{fig:pcc}
\end{figure*}

Both \cite{R20} and \cite{R19} introduced polyhedral conic classifiers that use PCFs and EPCFs with decision regions $f(\xx)\leq 0$ for positives and $f(\xx)>0$  for negatives (the opposite of the well known SVM - Support Vector Machine -  decision rule).
The resulting acceptance region includes an intersection of at most $2^d$ half spaces, and forms “kite-like" polyhedral acceptance regions as seen in Fig. \ref{fig:pcc}. The values of polyhedral conic functions given in (\ref{eq:con_func},\ref{eq:econ_func}) can be regarded as the distances to the acceptance regions in the sense that more negative values indicate that the test samples are closer to the center of the acceptance region and more positive values indicate that they are far from the acceptance region and therefore they must be rejected.

 \subsection{Deep Compact Polyhedral Conic Classifier}
\begin{figure}[tbh]
\centering
\begin{tabular}[h]{@{}cc@{}}
\includegraphics[width=0.5\columnwidth]{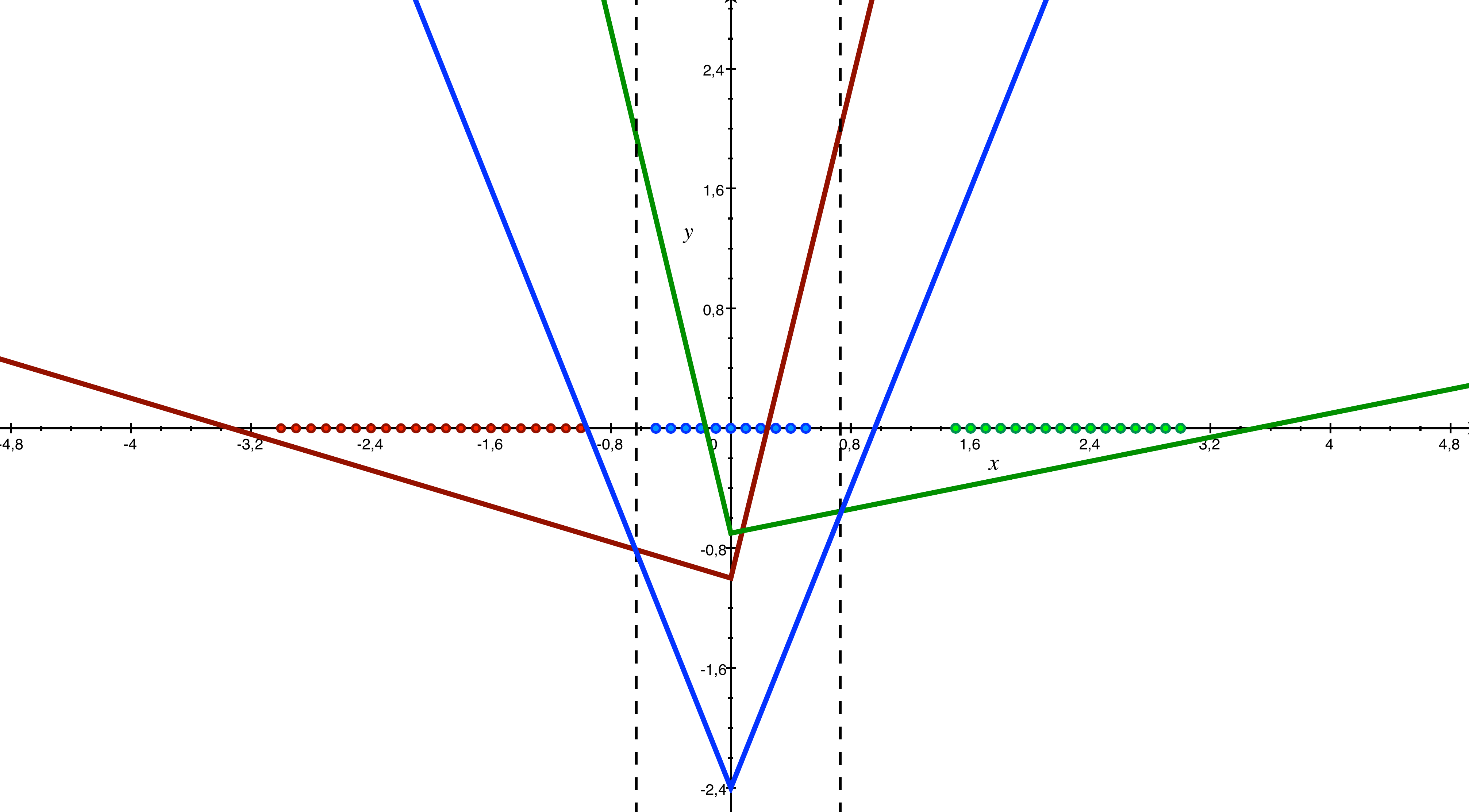} &
\includegraphics[width=0.5\columnwidth]{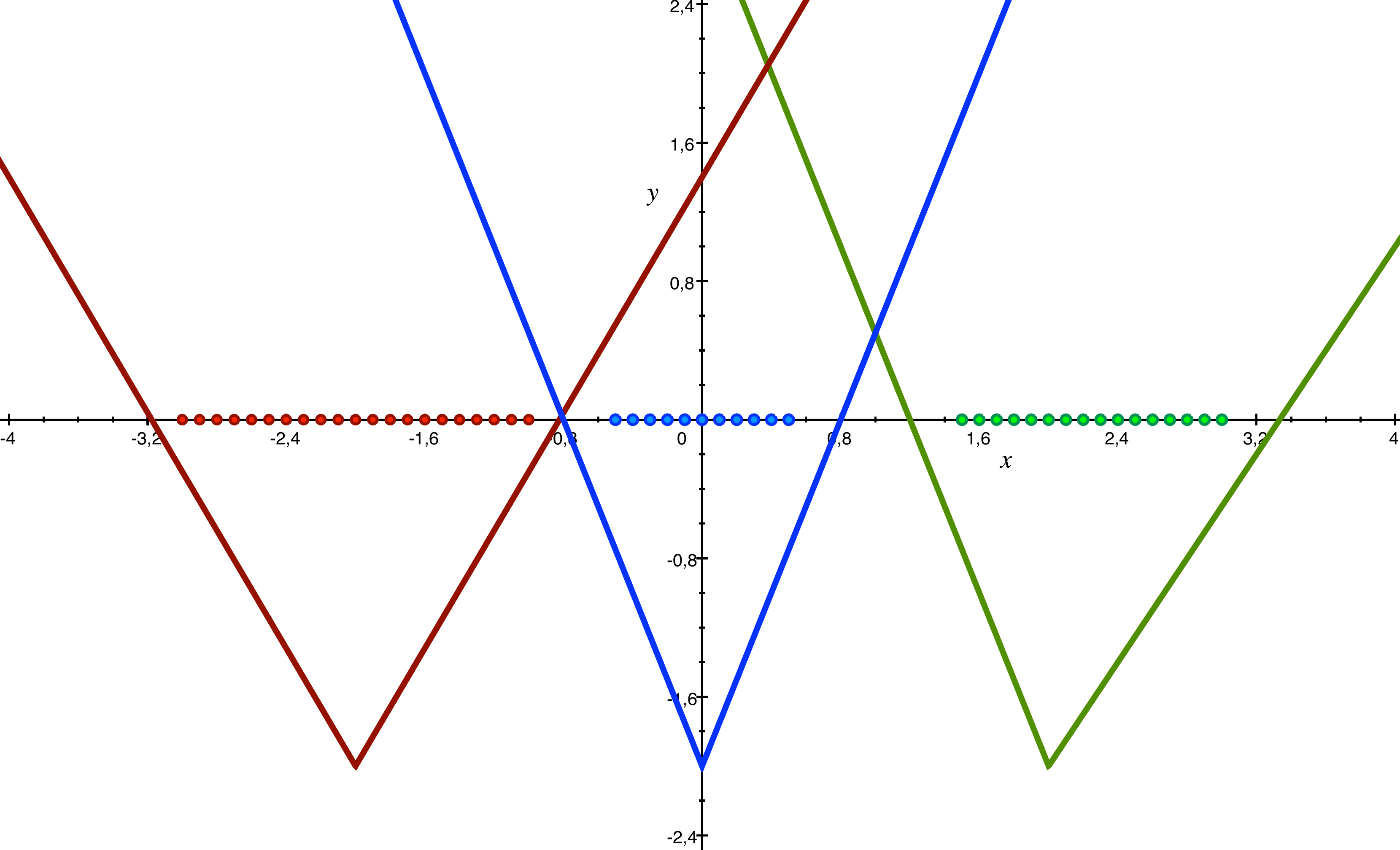}   \\
\end{tabular}
\caption{Using a single cone vertex for all three classes as seen on the left restrains to return better acceptance regions for classes. Using 3 cones (the separating cone of each class has the same color as the class samples) seen on the left figure still manages to return true acceptance regions determined by the dashed lines. However, this becomes more difficult or impossible for the classes whose samples lie far from the overall mean. In contrast, using a different cone vertex point $\cc_c$, $c=1,\ldots,3$, as seen on the right causes to return better acceptance regions easily. It also significantly improves the convergence time during training.}
\label{fig:2}
\end{figure}

Cevikalp and Saglamlar \cite{R19} introduced a deep neural network classifier using extended polyhedral conic function. However, this method possesses two major limitations: As a first limitation, instead of using a different cone vertex for each class, the authors used only a single common cone vertex for all classes in multi-class classification problems. This largely limits the classifier to return more discriminative class decision boundaries as illustrated in Fig. \ref{fig:2}. As seen in the figure, allowing only one common cone vertex for all classes, restrains the classifier to return well localized decision boundaries for the classes whose samples lie far from the selected cone vertex. This is because, the resulting acceptance region is formed by intersecting the cone (whose facets are learned by the classification algorithm) including the vertex $(\cc,-b)$ with the hyperplane as seen in Fig. \ref{fig:2}. Therefore, it is crucial to use a class-specific cone vertex point $\cc$ close to the class samples for each class. This also speeds up the training phase despite it brings a slight additional complexity during testing time. As a second limitation, \cite{R19} did not use any mechanism to return more compact class acceptance regions, which corresponds to minimizing the intra-class variations. Therefore, although the method took the inter-class separation into consideration, it neglected the intra-class variation as in soft-max loss function.

In this paper, we propose a new deep neural network classifier that simultaneously maximizes the inter-class margin and minimizes the intra-class variations. To this end, we first extend the method proposed in \cite{R19} such that different cone vertices are used for each class. Secondly, we introduce an elegant mechanism to adjust the compactness of the class acceptance regions, which in turn enables us to minimize the intra-class variations. 

It should be noted that the acceptance regions of the classes are bounded and convex only when $b>0$, $\xgamma>0$, and $|w_i|<\gamma_i$ for all $i=1,\ldots,d$,  i.e., when the hyperplane section has a shallower slope than every facet of the $L_1$ cone (see the proof given at the supplementary material).
Considering that these conditions are satisfied, the hyper-volume of the class acceptance regions depends on two factors: $\gamma_i$ values and $b$. For fixed $\gamma_i$ values, increasing the $b$ value increases the volume since the cone vertex is located at $(\cc,-b)$. This corresponds to move the $L_1$ cone downward without any change in its facets. Decreasing the $b$ value on the other hand will decrease the volume since this corresponds to move the cone upward closer to the intersecting hyperplane (see the related video given as supplementary material showing the volume change based on $b$ values). For a fixed $b$ value, we can also change the volume by changing the slopes of the rays forming the cone through $\gamma_i$ values. Increasing the $\gamma_i$ values decreases the volume of the acceptance region, whereas the decreasing $\gamma_i$ values increases the volume (see the related video).  
Therefore, the overall volume of the class acceptance regions is related to $b/\gamma_i$ since the acceptance region has width $O(b/\gamma_i)$ along axis $i$. Therefore, for fixed $b$ values, if we keep increasing $\gamma_i$ values, the acceptance region of a specific class shrinks and all corresponding class samples gather around a compact region. This effect is similar to the center loss term that enforces the samples of a specific class to cluster around their mean. In fact as seen in the synthetic experiments, the class samples typically collapse onto a single point in the proposed method although we do not explicitly enforce this constraint as in center loss function.

Now, assume we are given labeled training samples in the form $(\xx_i,y_i)$, $i=1,\ldots,n$ and $y_i \in \left\{1,\ldots,C\right\}$, where $C$ is the number of classes. The input samples $\xx_i$ can be 2D images or some vectoral inputs.
Let $\ff_i\in \R^d$ is the deep neural network feature representation of $\xx_i$ at the classification layer of the used network.
Consider that $\ww_c$ and $\xgamma_c$, $c=1,\ldots,C$, are the weights of the extended polyhedral conic classifier used in the last stage of the deep neural network.
By setting $\wn_c=-\ww_c$, $\gamman_c=-\xgamma_c$ (this is just for converting decision function into well-known SVM decision form),  the loss function of the proposed deep neural network method can be written as 
\begin{multline}
\label{eq:mopt}
\Scale[0.80]{\underset{\WWn,\bb}{\text{min}} \:\:\:\:\: \frac{\lambda}{2}\sum_{c=1}^C (\wn_c^\top \wn_c) + \sum_{i=1}^n \sum_{j\in \left\{1,...,C\right\}, j\neq y_i} H_1( (\wn_{y_i}^\top (\ff_i-\cc_{y_i}) }\\ 
\Scale[0.80]{+\gamman_{y_i}^\top  \left| \ff_i-\cc_{y_i} \right| + b_{y_i} ) - ( \wn_{j}^\top (\ff_i-\cc_{j})+\gamman_{j}^\top \left| \ff_i-\cc_{j} \right| + b_j ) )} \\
\Scale[0.80]{+ \eta \sum_{c=1}^C \sum_{m=1}^d \text{max}(0,\kappa-(-\gammani_{cm}-\left|\wni_{cm}\right|)).}
\end{multline}
Here, $H_\Delta(t) = \text{max}(0,\Delta-t)$ is the classical hinge loss, $\lambda$ is a regularization weight for $\wn_c$, $\WWn=\left[\wn_1 \ldots \wn_C \right]$, $\bb=\left[b_1 \ldots b_C \right]^\top$, and $\eta$ is the weight term for balancing the compactness loss term given as the last term with respect to other loss terms of the objective function. The $\kappa>0$ value given in the compactness term is a user-supplied cost penalty for ensuring $|w_i|<\gamma_i$ (or equivalently $-\gammani_i>|\wni_i|$) and decreasing values of $\gamman$ (increasing values of $\xgamma$) to minimize the hyper-volume of the class acceptance regions. The cone vertices, $\cc_{c}$, $c=1,\ldots,C$, are set to the mean of the classes. The first term in the optimization function is the regularization term for $\wn_c$ weights and it is already implemented in deep neural network classification methods as there are various methods to implement the regularization. The second term guarantees the large-margin separation and ensures that the classes are separated by a margin at least 1. The last term in the optimization controls the compactness of the class distribution, i.e., intra-class variation. To this end we use hinge loss again, and the positive $\kappa$ value acts as a margin to make sure that the positive $\gamma$ value is larger than $|w_i|$ by at least $\kappa$ in all directions, $m=1,\ldots,d$. The classes become more tight and compact as the value of $\kappa$ increases. This is similar to the center loss term in spirit, but we can adjust the compactness of the classes by changing the value of $\kappa$.

\begin{figure*}[t]
\centering
\begin{tabular}[tbh]{@{}cccc@{}}
\includegraphics[width=40mm]{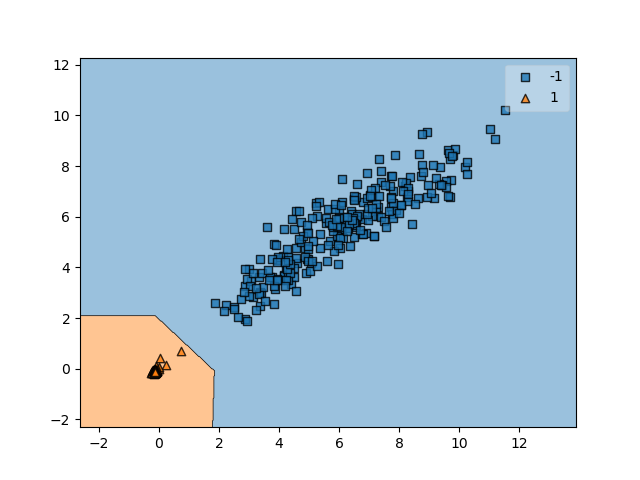} &
\includegraphics[width=40mm]{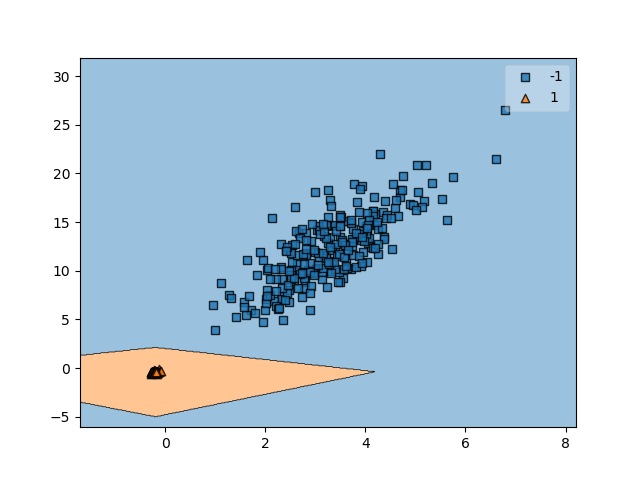}  & \includegraphics[width=40mm]{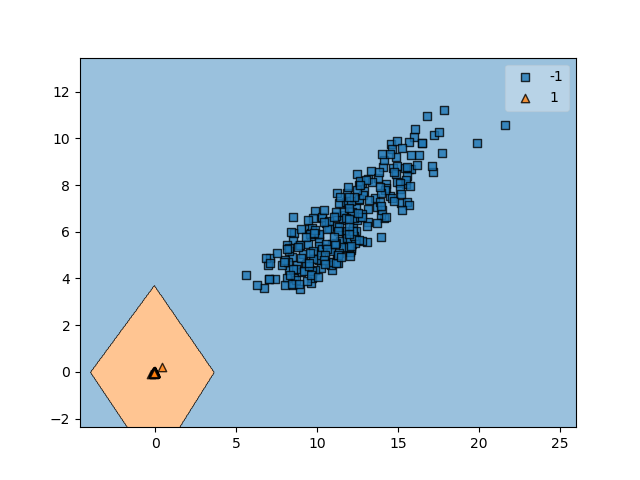} & \includegraphics[width=40mm]{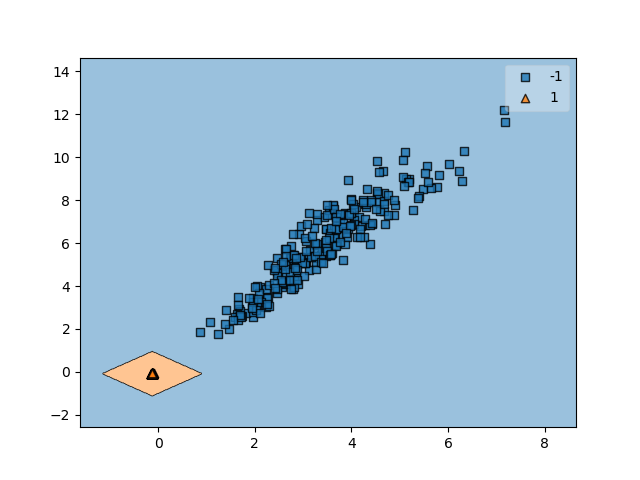} \\
$\kappa=0.0$ & $\kappa=0.35$ & $\kappa=0.70$ & $\kappa=1.0$
\end{tabular}
\caption{The positive class (shown with orange color) acceptance regions for different $\kappa$ values. If we do not enforce any constraint on the learned weights, i.e., $\kappa=0.0$, the acceptance region is unbounded. As we increase the value of $\kappa$, the positive class acceptance regions shrink, and the positive class samples tightly cluster around a compact region as expected. }
\label{fig:3}
\end{figure*}

Our proposed classification loss function allows to control both inter-class separation and intra-class variation, and it has a nice geometry that helps to interpret the resulting acceptance regions, which many recent methods lack. 
It should be noted that the Deep Support Vector Data Description (SVDD) method of \cite{R34} approximates a compact bounding hypersphere for a particular class. Compared to the polyhedral acceptance regions found by the proposed method, hyperspheres are looser models and they cannot approximate the class region more accurately compared to more tight bounding polyhedral regions returned by the proposed method.

\subsubsection{Implementation Details}
We used ResNet architecture \cite{R22} unless otherwise stated, and trained it with the SGD algorithm. It should be noted that the regularization term is already implemented in many deep neural network software, and it is commonly called as weight decay.
The common practice is to set weight decay to $0.00001$ for ResNet architectures, thus we tried values closer to this number. The best value is found to be equivalent to $0.0005$. Mini-batches of size $128$ were used in the experiments.
The momentum parameter was set to $0.9$, and we initialized the weights both randomly and by using the weights of a pre-trained network.
The $\kappa$ values control the compactness of the classes and we used validation data to set the best value.

We set the cone vertex of each class to the mean of the corresponding class samples. Since the feature representations change during training iterations, the class means need to be updated in each batch. However, updating the class means in each batch by using entire training set is impractical. Therefore, we update the class centers by using the selected samples in the current batch by using a similar procedure described in \cite{R1} (see supplementary material for details).

\section{Experiments}
We tested the proposed method, Deep Compact Extended Polyhedral Conic Classifier Network (DC-EPCC)\footnote{Source code of the proposed method is provided as supplementary material. The codes and trained models will be made publicly available.}, on both closed and open set recognition problems. We compared the proposed method to the Deep EPCC method of \cite{R19} and other related methods using soft-max loss and center loss functions \cite{R1,R2} as well as SphereFace \cite{R14}, CosFace \cite{R16}, and ArcFace \cite{R18} methods. We used ResNet-50 architecture as backbone architecture for all methods unless otherwise stated, thus they are all directly comparable. 

\subsection{Illustrations on Mnist and FaceScrub Datasets}
For the first experiment, we designed a deep neural network where the output of the last hidden layer is set to 2 for visualizing the learned features as in \cite{R1}. This allows to plot the learned feature vectors in 2D space. To visualize the control over the compactness of the classes by changing $\kappa$ values, we selected one class (the digit 2) from the Mnist dataset, and trained a binary deep neural network where the positive class is the selected digit class and the negative class is the remaining classes in the dataset. The learned feature representations of the positive class samples (shown with orange color) and the corresponding acceptance regions for different $\kappa$ values are plotted in Fig. \ref{fig:3}. As seen in the figure, if we do not enforce the compactness constraints on the learned weights (i.e., $\kappa=0.0$), the resulting positive class acceptance region is not necessarily bounded. As we increase the value of the $\kappa$, the area of the acceptance regions  decreases, which in turn yields more compact positive class acceptance regions. 
Moreover, although we do not explicitly enforce the class samples to collapse onto a single point as in center loss, positive class samples mostly collapse onto a single point, which is the ultimate goal in both center loss \cite{R1} and \textit{metric learning by collapsing classes} method introduced in \cite{R31}.

\begin{figure*}[tbh]
\centering
\begin{tabular}[h]{@{}ccc@{}}
\includegraphics[width=42mm]{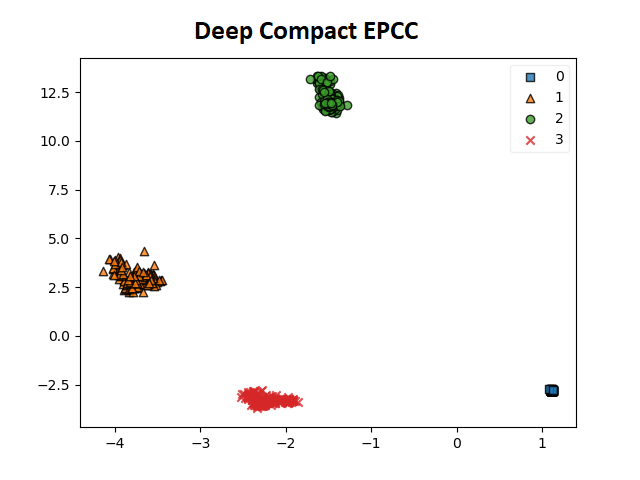} &
\includegraphics[width=42mm]{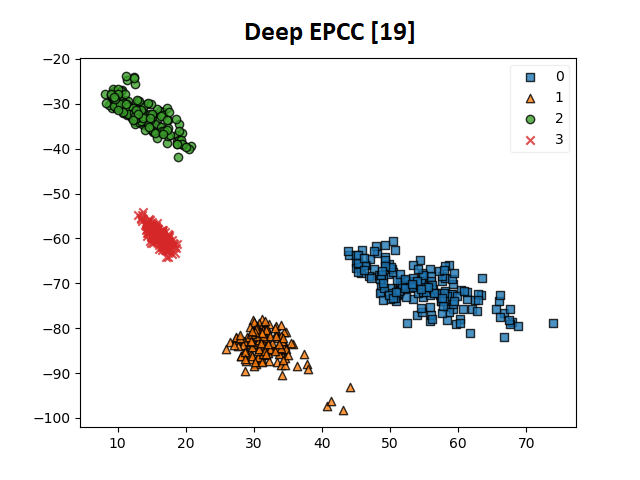}  & \includegraphics[width=42mm]{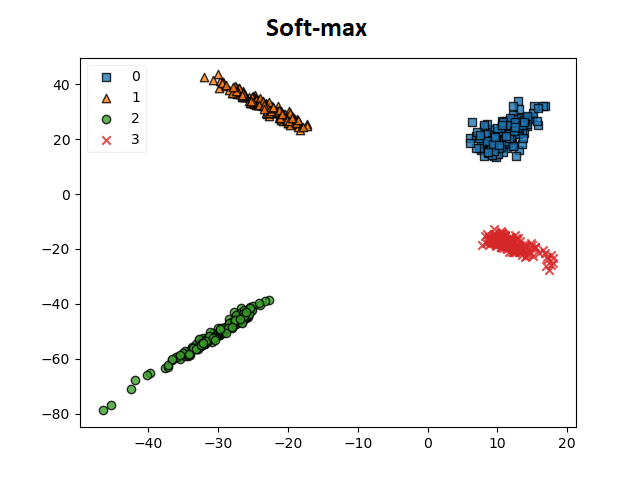}  \\
\includegraphics[width=42mm]{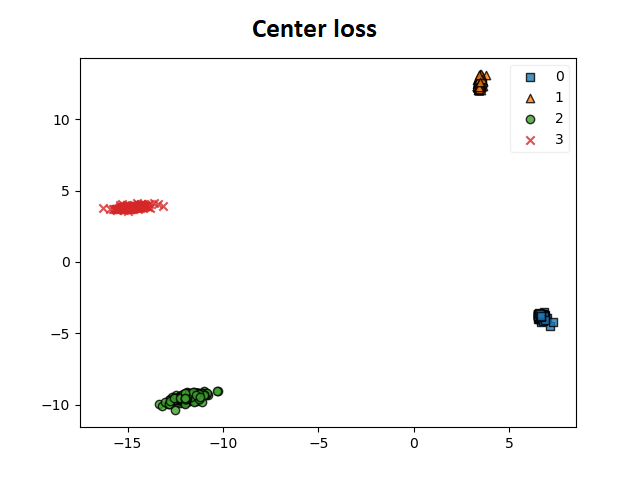} & 
\includegraphics[width=42mm]{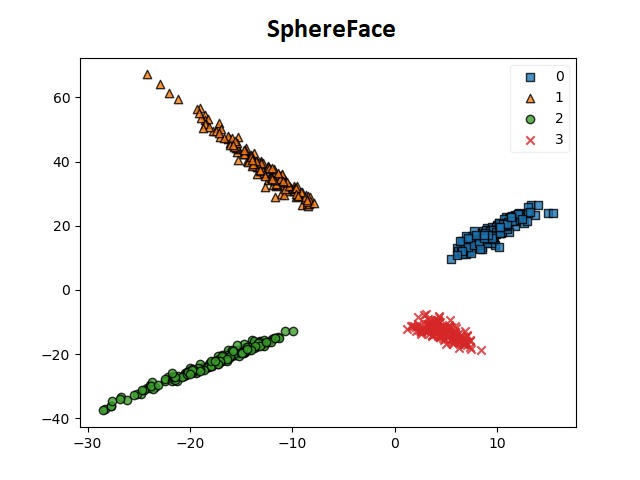} & \includegraphics[width=42mm]{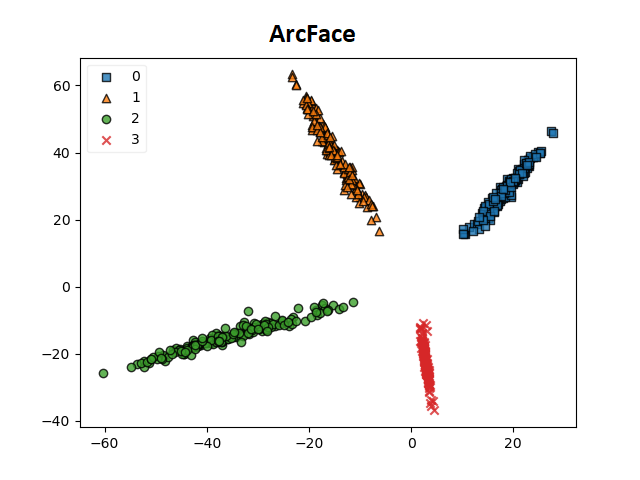}
\end{tabular}
\caption{Learned feature representations of face images for different methods. }
\label{fig:4}
\end{figure*}

In the second experiment, we compared the learned feature representations returned by various deep neural networks including the proposed method. To this end, we used FaceScrub dataset. The FaceScrub dataset \cite{R23} includes face images of 530 celebrities. It has been created by detecting faces based on automated search of public figures on the internet followed by manually checking and cleaning the results. In the dataset, there are 265 male and 265 female celebrities’ face images. In our experiments, we used only 4 classes for better visualization. The learned features are plotted in Fig. \ref{fig:4}. We also plotted the feature representations learned by the deep EPCC method of \cite{R19} that uses a common center for all classes and does not enforce any compactness criterion. As seen in the figure, both the proposed method and the method using center loss return compactly clustered feature representations, whereas deep EPCC and soft-max loss fail to return compact class regions. The SphereFace and ArcFace methods enforce compactness in the angular space, thus the features learned by these two methods lie in the vicinity of lines with different orientations. Therefore, the angular orientations of these lines determine the class memberships during testing. We would like to point out that the proposed method returns more compact feature representations with respect to deep EPCC method of \cite{R19} as expected.

\subsection{Experiments on PASCAL VOC 2007 Dataset}
We conducted experiments on the PASCAL 2007 Visual Object Classification dataset. PASCAL VOC dataset includes 20 object categories. We trained a binary classifier for each category where the object class is taken as positive and the remaining classes are treated as negatives, which is the standard classification setup. 
The accuracies are computed in terms of average precision by using the VOC evaluation toolkit. 

The results are given in Table \ref{tb:pascalvoc}. SphereFace, CosFace and ArcFace method yielded very poor accuracies, thus we omitted these results. As seen in the table, the proposed DC-EPCC method significantly outperforms CNN networks using both soft-max and center loss functions. The proposed method also significantly outperforms Deep EPCC method of \cite{R19}. Our proposed method achieves the best accuracies for 15 categories of all 20 object categories, and it outperforms the second best performing method by 2\% on the average. The CNN method using the center loss wins for the remaining 5 object categories, and the CNN network using the soft-max is the worst performing method. Deep EPCC method of \cite{R19} also outperforms the CNN method using soft-max loss, but its accuracy is lower than the accuracies of the proposed method and the method using center loss. It should be noted that both deep EPCC methods use positive class center in this setting, thus the only difference between the two methods is that we enforce the class compactness in the proposed methodology. Yet, the accuracy difference is quite big, 3.7\%, which shows the importance of compact acceptance regions.
Obtaining best accuracies by the proposed method and center loss is mostly expected since both methods return compact class acceptance regions for the tested positive object classes and they easily reject the samples coming from the background classes. However, a complete failure of the methods maximizing margin in the angular space was quite unexpected.

\begin{table*} [tb] 
	\begin{center}
		\renewcommand{\arraystretch}{1.2}
		\caption{Average Precision scores (\%) on PASCAL VOC 2007 classification dataset.}
		\label{tb:pascalvoc}
		\def\S#1{\begin{sideways}#1\end{sideways}}
		\def\B#1{\textbf{#1}}
		\small
		\tabcolsep 2.2pt   
		\resizebox{\textwidth}{!}{
		\begin{tabular} {|@{~\strut}l@{~}*{21}{|c}|} \hline
			\B{Methods}  & 
			\S{Aeroplane}&\S{Bicycle}&\S{Bird}&\S{Boat}&\S{Bottle}&\S{Bus}&\S{Car}&\S{Cat}&\S{Chair}&\S{Cow}&\S{Dining Table~~}&\S{Dog}&\S{Horse}&\S{Motorbike}&\S{Person}&\S{Potted Plant}&\S{Sheep}&\S{Sofa}&\S{Train}& \S{TV Monitor} & \S{Average}\\ \hline 
			{DC-EPCC} & 93.9 & \textbf{89.2} & \textbf{90.5} & 90.2 & \textbf{51.9} & \textbf{82.5} & \textbf{91.1} & 87.8 & \textbf{69.5} & \textbf{77.0} & \textbf{75.7} & \textbf{85.0} & \textbf{91.8} & \textbf{86.8}  & 94.2 & \textbf{61.4} & \textbf{81.2} & \textbf{73.1} & 94.7 & \textbf{82.4} & \textbf{82.5} \\ \hline
			{Deep EPCC \cite{R19}} & 93.2 & 85.6 & 86.3 & 87.4 & 44.1 & 79.9 & 88.0 & 84.6 & 66.9 & 69.5 & 72.0 & 79.8 & 91.6 & 85.0  & 94.1 & 58.8 & 79.2 & 67.5 & 94.4 & 73.1 & 78.8 \\ \hline
			{Soft-max} & 91.9 & 85.0 & 84.7 & 86.1 & 40.9 & 73.5 & 88.3 & 80.3 & 65.0 & 60.9 & 64.5 & 75.6 & 88.4 & 81.0 & 91.4 & 49.3 & 66.7 & 63.3 & 90.7 & 72.0 & 75.0 \\ \hline
			{Center Loss \cite{R1}} & \textbf{94.3} & 88.0 & 88.0 & \textbf{90.9} & 48.0 & 78.9 & 90.7 & \textbf{88.3} & 68.5 & 71.1 & 74.1 & 84.2 & 90.8 & 85.7 & \textbf{94.5} & 57.0 & 75.3 & 68.4 & \textbf{94.8} & 78.4 & 80.5 \\ \hline
			\end{tabular}
			}
	\end{center}
\end{table*}

\subsection{Experiments on CIFAR-100 and FaceScrub Datasets}
Here we tested the proposed method on CIFAR-100 and FaceScrub datasets. We fine-tuned all tested networks from a network trained on ILSVRC2012 for CIFAR-100 dataset experiment and a network trained on MS-Celeb-1M dataset \cite{R32} is used to fine the CNN networks for FaceScrub dataset experiment. The accuracies of methods tested on CIFAR-100 dataset are given in terms of classical classification rates in Table \ref{cifar100_res}. The proposed DC-EPCC method achieves the best accuracy followed by the CNN method using center loss. Methods maximizing the margin in the angular space, SphereFace, CosFace and ArcFace, perform badly and give the worst accuracies. This shows that general object categories cannot be modeled by subspaces as in face recognition. Deep EPCC method of \cite{R19} also performs poorly mostly due to the fact that using a common single center for all 100 classes restricts to return well localized acceptance regions as illustrated in Fig. \ref{fig:2}.

\begin{table}[t]
	\begin{center}
		\renewcommand{\arraystretch}{1.3}
		\caption{Classification Rates (\%) on the CIFAR-100 and FaceScrub datasets.}
		\label{cifar100_res}
		\vspace{0.3cm}
		\begin{tabular}[tb] {|l|c|c|} \hline
			\textbf{Method}  & \textbf{CIFAR-100}  & \textbf{FaceScrub}\\ \hline
			DC-EPCC & \textbf{83.17} &	98.19 \\ \hline
			Deep EPCC \cite{R19} & 78.06 & 98.18  \\ \hline
			Soft-max & 81.88 & 96.02 \\ \hline
			Center Loss \cite{R1} & 82.19 & \textbf{98.99} \\ \hline
			SphereFace \cite{R14} & 75.19 & 97.50 \\ \hline
			CosFace \cite{R16} & 78.85 & 97.97\\ \hline
			ArcFace \cite{R18} &  77.74 & 97.65 \\ \hline
		\end{tabular}
	\end{center}
	\vskip-10pt
\end{table}

The FaceScrub dataset \cite{R23} includes face images of 530 celebrities. In this dataset, there are 265 male and 265 female celebrities' face images.
We manually checked the face images and cleaned non-face images since there were still some annotation mistakes. As a result, we had 67,437 face images of 530 celebrities with an average of 127 images (minimum 39, maximum 201) per person. The face images are mostly high resolution frontal face images and we resized them to 112$\times$112. 
We used 60\% of data as training, and the remaining 40\% is used for testing. The classification accuracies of tested methods are given 
in Table \ref{cifar100_res}. The best accuracy is obtained by CNN method using center loss and it is closely followed by the proposed method. As opposed to CIFAR-100 results, methods maximizing the angular margin also perform well here. The CNN method using the soft-max is the worst performing method.

\subsection{Experimental Results on Open Set Recognition}
For open set recognition, we follow the settings used in \cite{R38,R39} and evaluate open set recognition performance on SVHN \cite{R42} and CIFAR-10 datasets. In this setting, samples selected from randomly chosen 6 classes are used for training, and test set uses the samples coming from all 10 classes. The goal is to detect the novel samples coming from the unknown 4 classes during testing. This procedure is repeated 5 times and average of the computed accuracies in each run is used for final accuracy. Area under the ROC (Receiver Operating Characteristic) curve (AU-ROC) is used to evaluate the performance. In the proposed method, we first scale classifiers' scores by dividing the maximum of the absolute values for each class such that scores for a particular class will be between -1 and 1 (we also applied sigmoid function to convert the scores to probabilities, the results were similar). Then, the negative of the maximum score is used as the final score value for open-set recognition, i.e., $-\underset {c=1,...,C} {\text{max}} \: s_c(\ff_{test})$, where $s_c(\ff_{test})$ denotes the classification score of the test sample for the class $c$.

\begin{table}[t]
	\begin{center}
		\renewcommand{\arraystretch}{1.3}
		\caption{AU-ROC Scores (\%) for open-set recognition. Results are averaged over 5 random partitions of known/open set classes. All the values in the table other than the proposed method and Deep EPCC are taken from \cite{R38} and \cite{R39}.}
		\label{openset_res}
		\vspace{0.3cm}
		\begin{tabular}[tb] {|l|c|c|} \hline
			\textbf{Method}  & \textbf{CIFAR-10}  & \textbf{SVHN} \\ \hline 
			DC-EPCC & 87.7 &	\textbf{93.2} \\ \hline 
			Deep EPCC \cite{R19} & 84.8 & 92.0\\ \hline 
			Soft-max & 67.7 &  88.6 \\ \hline 
			OpenMax \cite{R33} &	69.5 & 89.4\\ \hline 
			G-OpenMax \cite{R41} &	67.5 & 89.6 \\ \hline 
			OSRCI \cite{R38}  &	69.9 & 91.0 \\ \hline 
			C2AE \cite{R39} & \textbf{89.5}  & 92.2 \\ \hline 
		\end{tabular}
	\end{center}
	\vskip-10pt
\end{table}

Computed accuracies are given in Table \ref{openset_res}. All the values in the table other than the proposed method and Deep EPCC are taken from \cite{R38} and \cite{R39}. As seen in the table, the proposed method achieves the best accuracy for SVHN dataset and obtains a comparable result on the CIFAR-10 dataset. The results on CIFAR-10 datasets are very promising in the sense that the proposed method and C2AE significantly outperform other open set recognition methods.
It should be noted that, we do not attempt to do any special operation for the open set recognition such as creating synthetic open set examples as in \cite{R38} and \cite{R33}. We train our classifiers as usual as in closed set recognition settings. Yet, we significantly outperform many of the state-of-the-art open set recognition algorithms as seen in the table.
Obtaining good accuracies by the proposed method is expected since the proposed method returns very compact acceptance regions for classes. The samples in the center of these regions have the higher confidence scores, and the scores decrease as the samples move away from the center. Therefore, the samples coming from different classes lie far from these acceptance regions and they have very low confidence scores. As a result, they are easily rejected, which in turn improves the AU-ROC score.

\begin{table}[t!]
	\begin{center}
		\renewcommand{\arraystretch}{1.2}
		\caption{Performance Comparison of various approaches for different views and modalities on the DAD dataset.}
		\label{dad_res}
		\tabcolsep 1.5pt  
		\begin{tabular}[tb] {|l|c|c|c|c|} \hline
			\multicolumn{1}{|c|}{\multirow{2}{*}{\textbf{Method}}} & \multicolumn{4}{c|}{\textbf{AU-ROC}} \\  \cline{2-5}   
			& \phantom{i} \textbf{Top-D} \phantom{i}  &\phantom{i} \textbf{Top-IR} \phantom{i}& \phantom{i} \textbf{Front-D} \phantom{i} & \textbf{Front-IR} \\ \hline 
			DC-EPCC 			  & 0.9012  & \textbf{0.8954}  & \textbf{0.9087}  & \textbf{0.8727} \\ \hline 
			Deep EPCC \cite{R19}  & 0.8859  & 0.8665  & 0.8832  & 0.8715 \\ \hline 
			CL \cite{R43}      & \textbf{0.9200}  & 0.8857  & 0.9020  & 0.8666 \\ \hline
			Deep SVDD \cite{R34}  &	0.5453  & 0.5881  & 0.6509  & 0.6434 \\ \hline
		\end{tabular}
	\end{center}
	\vskip-10pt
\end{table}

\subsection{Experiments on Anomaly Detection Task}
For anomaly detection task, we use recently introduced Driver Anomaly Detection (DAD) dataset \cite{R43}. The DAD dataset consists of 550 minutes of normal driving videos and 100 minutes of anomalous driving containing 8 different distracting actions in its training set. In the test set of the DAD dataset, there are 88 minutes of normal driving videos and 45 minutes of anomalous driving videos containing 24 different distracting actions. The dataset is recorded with 45 fps for two views (front and top) and two modalities (depth and infrared). The task here is to generalize normal driving and detect anomalies although some of the distracting actions has never been introduced to the network at the training.

We have compared our approach with several anomaly detection methods. In order to make fair comparison, all the models use 3D ResNet-18 network that is pretrained on Kinetics-600 dataset. All methods receive 16 frames input downsampled from 32 frames. Comparative results can be seen in Table \ref{dad_res}. The proposed method achieves the best performance for front view with both modalities and top view with infrared modalities. For only top view depth modality, contrastive learning approach used in \cite{R43} achieves the best performance. We must also note that Deep SVDD \cite{R34} method performs significantly inferior compared to other methods. The reason is that this method only uses normal driving videos at the training time and does not leverage anomalous driving samples in the training set (although the classical SVDD method of \cite{R35} allows the negative samples for learning, the deep neural network version does not have this property). This shows the importance of the anomalous samples at the training time to generalize normal driving.

\begin{figure}[t]
	\centering
	\includegraphics[width=85mm]{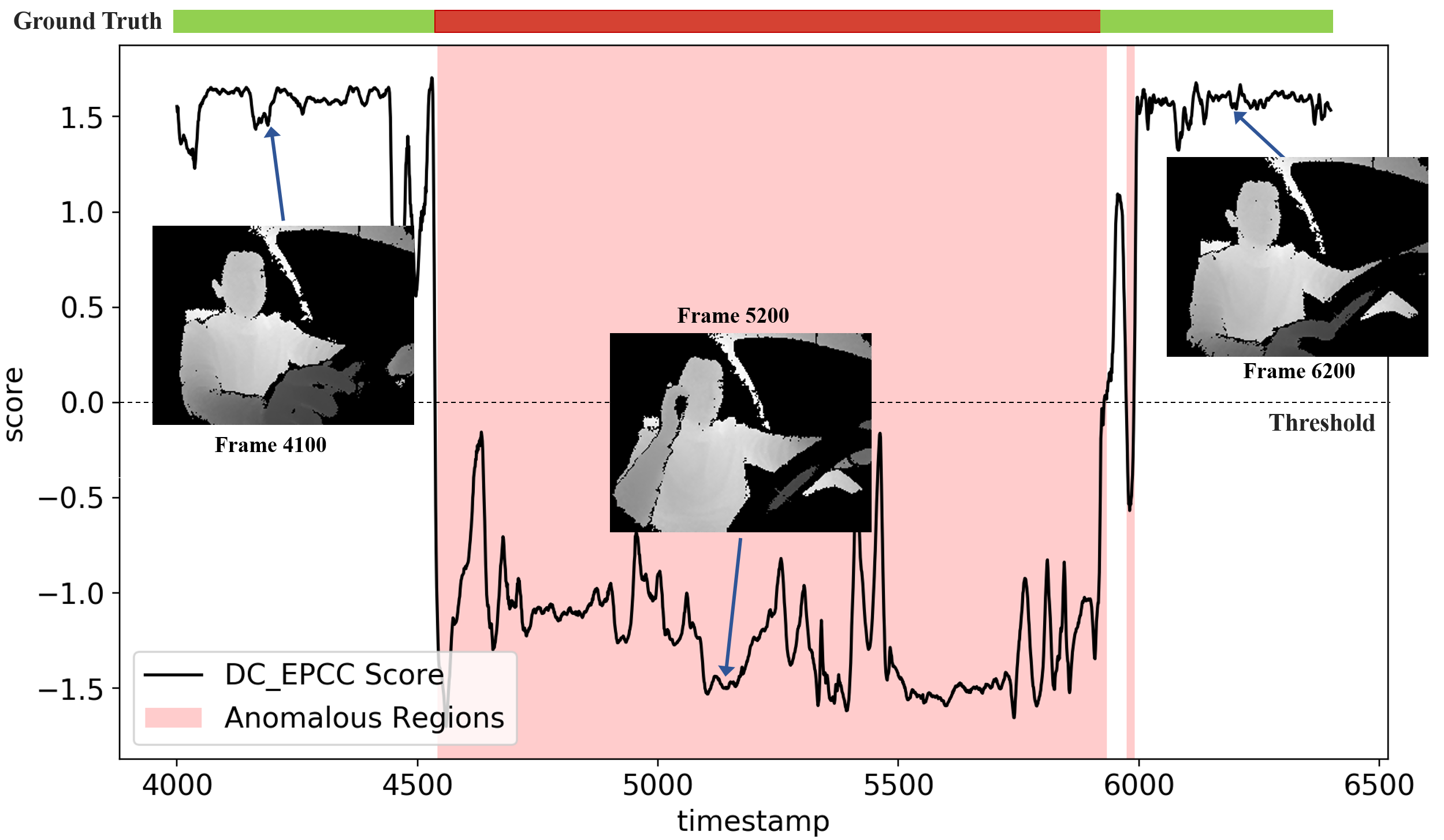}
	\caption{Illustration of recognition for a continuous video stream using front view and dept modality. The frames are classified as anomalous driving if the similarity score is blow the preset threshold. Ground truth annotation is shown on top.}
	\label{fig:DAD_scores}
\end{figure}

We have also visualized the generalization perfromance of the proposed DC-EPCC method on a continuous video stream from the DAD dataset validation set in Fig. \ref{fig:DAD_scores}. DC-EPCC can successfully produce high scores for normal driving and low scores for anomalous actions. Therefore, utilization of a preset threshold is enough for the detection of anomalies. Although there are some fluctuations at the transition from normal to anomalous (and vice versa), they can be mitigated by applying some post processing similar to \cite{R43}. 

\section{Conclusion}
This paper introduces a new deep neural network classifier that simultaneously minimizes the intra-class variations and maximizes the inter-class separability. To this end, the proposed method uses a margin based cost function for maximizing the inter-class separation, and a geometrically inspired mechanism that controls the compactness of the class acceptance regions to minimize the intra-class variation. The experimental results on various datasets show that the proposed method outperforms related state-of-the-art deep neural network classifiers on majority of the tested datasets. Especially, the proposed classifier is more suitable for open set recognition tasks and binary classification problems in which one positive object class is separated from a background class composed of many classes other than the object of interest as in PASCAL VOC classification setting.
Moreover, the proposed method achieves the state-of-the-art accuracies on anomaly detection and significantly outperforms state-of-the-art method, Deep SVDD, proposed for anomaly detection.

{\small
\bibliographystyle{ieee_fullname}

}

\end{document}